\documentclass{isprs}
\usepackage{subfigure}
\usepackage{setspace}
\usepackage{geometry} 
\usepackage{enumerate}
\usepackage{amsmath}
\usepackage{amssymb}
\usepackage{natbib}
\usepackage{multirow}  
\usepackage{stfloats}
\usepackage{caption}
\usepackage{microtype}
\usepackage{hyperref}

\usepackage{graphics}
\graphicspath{figures/}
\begin{document}

\title{Density-Aware Convolutional Networks with Context Encoding for Airborne LiDAR Point Cloud Classification}

\author{
Xiang Li\textsuperscript{a,b}, Mingyang Wang\textsuperscript{a,b}, Congcong Wen\textsuperscript{b}, Lingjing Wang\textsuperscript{a,b}, Nan Zhou \textsuperscript{c}, Yi Fang\textsuperscript{a,b}\thanks{Corresponding author. Email: yfang@nyu.edu.}
}
\address
{
\textsuperscript{a } Multimedia and Visual Computing Lab, New York University, New York, United States.\\
\textsuperscript{b } Tandon School of Engineering, New York University, New York, United States.\\
\textsuperscript{c } Aerospace Information Research Institute, Chinese Academy of Sciences.\\
}

\abstract
{
To better address challenging issues of the irregularity and inhomogeneity inherently present in 3D point clouds, researchers have shifted their focus from designing hand-crafted point features toward learning 3D point signatures using deep neural networks for 3D point cloud classification. Recently proposed deep learning-based point cloud classification methods either apply 2D CNNs on projected feature images or apply 1D convolutional layers directly on raw point sets. These methods cannot adequately recognize fine-grained local structures caused by the uneven density distribution of the point cloud data. In this paper, to address this challenging issue, we introduced a density-aware convolutional module that uses the pointwise density to reweight the learnable weights of convolution kernels. The proposed convolution module can fully approximate the 3D continuous convolution on unevenly distributed 3D point sets. Based on this convolution module, we further developed a multiscale fully convolutional neural network with downsampling and upsampling blocks to enable hierarchical point feature learning. In addition, to regularize the global semantic context, we implemented a context encoding module to predict a global context encoding and formulated a context encoding regularizer to enforce the predicted context encoding to align with the ground truth encoding. The overall network can be trained in an end-to-end fashion with the raw 3D coordinates and the height above ground as inputs. Experiments on the International Society for Photogrammetry and Remote Sensing (ISPRS) 3D labeling benchmark demonstrated the superiority of the proposed method for point cloud classification. Our model achieved a new state-of-the-art performance with an average F1 score of 71.2\% and improved the performance by a large margin on several categories (such as powerline, impervious surface, car, and facade).

}

\keywords{Airborne LiDAR, Point cloud classification, Density-aware convolution, Context encoding}

\maketitle

\section{Introduction}\label{Introduction}
With the rapid development of 3D sensor technology, 3D point cloud data have become increasingly accessible through innovations in light detection and ranging (LiDAR), synthetic aperture radar (SAR), dense stereo- or multiview-photogrammetry in remote sensing and computer vision fields. Among these techniques, airborne light detection and ranging (LiDAR), also known as laser scanning, provides reliable 3D spatial information and plays an important role in many applications, such as topographic mapping, forest monitoring \citep{axelsson2000generation,mongus2013computationally}, power line detection \citep{andersen2005estimating,solberg2009mapping,zhao2009lidar,ene2017large}, road detection and planning, and 3D building reconstruction \citep{kada20093d,yang2017automated}. Despite the prevalence of 3D point cloud data, automatic classification and segmentation of 3D point clouds remain challenging due to the irregular structure of raw point clouds.

To classify the point clouds, early efforts mostly focused on either the design of geometric features to characterize the local structure of each point, such as density, curvature, roughness, or the development of discriminative models, e.g., Gaussian mixture model \citep{lalonde2005scale,lalonde2006natural}, support vector machine \citep{zhang2013svm}, AdaBoost \citep{lodha2007aerial}, and random forest \citep{babahajiani2017urban,chehata2009airborne}, for the classification task. Other studies have tried to boost the performance by incorporating contextual information to enforce label consistency \cite{munoz2009contextual, shapovalov2010nonassociative,niemeyer2014contextual,weinmann2015contextual,niemeyer2012conditional}. However, these contextual-based approaches still employ hand-crafted features, and thus, fail to adequately extract high-level semantic structures, and the generalization ability of these models was limited when applied to large-scale wild scenes.

In recent years, with the prevalence of deep learning methods in remote sensing fields, remarkable performance has been achieved in various applications, including scene classification, object detection, change detection, and hyperspectral image classification. \cite{hu2015transferring,hu2015deep,maggiori2016convolutional,cheng2016learning,zhan2017change,li2018building}. Given the great success of deep learning-based methods for remote sensing image recognition, researchers have been shifting their focus toward deep learning-based methods for 3D point cloud classification \citep{qi2017pointnet,li2018pointcnn,yang2017convolutional,yousefhussien2018multi,wang2018deep}. For example, to make use of the great power of 2D convolutional neural network (CNN) for image recognition, some researchers \citep{yang2017convolutional,zhao2018classifying} proposed to project 3D point clouds into 2D feature images and then employ conventional 2D CNN for airborne point cloud classification. These methods usually need to calculate additional hand-crafted features to enrich the 2D feature image representations, and classification performance is limited due to the information loss during 3D to 2D transformation. More recent works have attempted to directly apply convolutions on irregular point clouds for the classification task \citep{yousefhussien2018multi,wang2018deep,wen2019directionally}. Although these methods have achieved state-of-the-art performance on several point cloud classification benchmarks, they do not adequately recognize fine-grained local structures due to the uneven density distribution of the point cloud data.

To address this issue, in this paper, we proposed a novel density-aware point convolution module to extract locally representative features of 3D point clouds. Our key innovation is to force the convolution module to be aware of the local density distribution when learning its kernel weights. To achieve the goal of 3D point cloud classification, we further developed a multiscale fully convolutional neural network with downsampling and upsampling blocks to enable hierarchical point feature learning and per-point label prediction. Moreover, considering the imbalanced class distribution of an outdoor scene, a context encoding module was integrated into our network to regularize the global semantic context. The overall network can be trained in an end-to-end fashion and directly predict the classification labels for all the input points in one forward pass. We list the main contributions of the proposed method as follows:

\begin{enumerate}[1.]
\item This paper introduces a novel density-aware convolution module that directly applies convolutions on irregular point clouds to learn representative point features.
\item With the proposed density-aware convolution module, we further develop a multiscale fully convolutional neural network with downsampling and upsampling blocks for the task of 3D point cloud classification.
\item We introduce a context encoding loss to regularize the global semantic context and experimentally demonstrate its effectiveness.
\item We eliminate the need for calculating costly hand-crafted features or corresponding spectral information and achieve superior performance on the ISPRS 3D labeling benchmark dataset.

\end{enumerate}

The remainder of this paper is organized as follows. In Section \ref{Related Work}, we provide a brief review of the airborne LiDAR point cloud classification methods. The proposed density-aware convolutional network is described in detail in Section \ref{Methods} In Section \ref{Experiments}, we conduct experiments to verify the classification performance of the proposed method. We further discuss the effectiveness of the proposed density-aware convolution module and context encoding module in Section \ref{sc_discussion} Finally, the paper concludes in Section \ref{sc_conclusion}

\section{Related Work}\label{Related Work}
Point cloud segmentation methods can be generally divided into two main categories: traditional nonlearning-based methods and deep learning-based methods. 

\subsection{Traditional Nonlearning-based Methods}\label{sc_tradition}
Traditional nonlearning-based methods start by designing hard-crafted point features and then employing simple pointwise discriminative models to label the input point sets. Such hand-crafted features are commonly generated from the covariance matrix of a local neighborhood and provide local geometry features of each point, e.g., planarity, sphericity, and linearity (Lin et al., 2014a). Subsequently, some conventional supervised learning algorithms, such as random forests (RFs), support vector machines (SVMs), Bayesian networks (BNs) and AdaBoost, are employed to learn discriminative models from a set of training samples. \citeauthor{lodha2007aerial} proposed to train an AdaBoost algorithm to classify airborne LiDAR point clouds into four categories (i.e., road, grass, buildings, and trees) based on five manually defined features including height, height variation, normal variation, intensity, and image intensity \citep{lodha2007aerial}. \citeauthor{kim2011random} adopted the random forest classifier for powerline scene classification based on two different sets of features extracted in the point domain and a feature (i.e., line and polygon) domain \citep{kim2011random}. Kang et al. (2016) first extracted geometric features from point clouds and spectral features from optical images and trained a BN classifier for airborne LiDAR point cloud classification. \citeauthor{zhang2013svm,mallet2011relevance} employed the support vector machine (SVM) algorithm to classify airborne LiDAR point clouds of the urban scene based on thirteen features of geometry, radiometry, topology and echo characteristics \citep{zhang2013svm,mallet2011relevance}. \citeauthor{chehata2009airborne} proposed to use the random forest algorithm for point cloud classification based on twelve hand-crafted features regarding optical, multiecho lidar and FW lidar components and analyzed the variable importance of each hand-crafted feature for the classification of urban scenes \citep{chehata2009airborne}.
However, these methods usually classify each point independently with its local features and neglect the semantic labels of its neighboring points, which can easily lead to classification noises and label inconsistency, especially in complex scenes\citep{weinmann2015semantic}.


To address this issue, recent studies have developed several context-based classification approaches to improve the smoothness of the classification results. For example, \citeauthor{niemeyer2014contextual} developed a contextual classification method based on conditional random fields (CRF). Their experimental results demonstrate an improvement of 2\% in the overall accuracy by integrating contextual features \citep{niemeyer2014contextual}. 
\citeauthor{niemeyer2016hierarchical} further enhanced contextual information by incorporating a two-layer conditional random field (CRF), where the first layer applies on the point level to generate segments and the second layer operates on the generated segments and incorporates a larger spatial scale \citep{niemeyer2016hierarchical}. \citeauthor{garcia2015evolutionary} proposed a contextual classification method based on a support vector machine (SVM) and an evolutionary majority voting technique. The proposed method achieved superior performance for land cover classification compared to pixel-based SVM as well as a contextual classified based on SMV and MRF \citep{garcia2015evolutionary}. \citeauthor{munoz2009contextual} proposed to incorporate the associative Markov network (AMN) to enable better high-order contextual interactions for the classification of 3D LiDAR point clouds classification \citep{munoz2009contextual}. \citeauthor{shapovalov2010nonassociative} further proposed the nonassociative Markov networks model by using dynamic instead of constant pairwise potentials in AMN for a pair of different class labels.

However, the above nonlearning-based methods need to manually extract pointwise local and contextual features in advance. This involves unwanted preprocessing time and makes these models sensitive to the quality of feature engineering. Moreover, their classification performance degrades when dealing with the point clouds scanned from complex scenes \citep{zhao2018classifying}.

\subsection{Deep learning-based methods}\label{sc_learning}
In contrast to the above nonlearning-based methods, deep learning-based methods can automatically extract high-level features from a large quantity of input data without a hierarchical deep neural network.
In general, these deep learning-based point cloud classification methods can be grouped into two main categories: feature image-based methods and point-based methods.

\subsubsection{Feature image-based methods}~\\
Convolutional neural networks (CNNs), one of the commonly used deep learning methods, have achieved some extremely promising results in various 2D image recognition tasks, such as scene classification, object detection, and semantic segmentation. However, the unordered and irregular nature of 3D point clouds poses great challenges for the direct extension of 2D CNNs. Therefore, early researchers tried to transform the 3D point cloud into more tractable 2D images and then employ convolutional 2D CNN to achieve point cloud classification and segmentation \citep{yang2017convolutional,yang2017convolutional,yang2018segmentation,zhao2018classifying}. For example, \cite{su2015multi} proposed to first generate multiple 2D rendered images of 3D shapes, and then a conventional 2D CNN was used to extract features of each view. A view-pooling (aggregation) layer was further proposed to fuse information from multiple views and boost the classification performance. A similar method was proposed in \citeauthor{yang2017convolutional}, which generated projected 2D feature images that characterized the local geometric features, global geometric features and full-waveform features of each point.

For airborne LiDAR point clouds, \citeauthor{yang2018segmentation} developed a protective convolutional neural network-based method for airborne LiDAR point cloud classification. Their method starts by transforming 3D point clouds into 2D feature images that characterize the height, intensity, planarity, sphericity, and variance in deviation angles across multiple scales. Then, a 2D CNN is developed to learn representative features from multiple 2D projected feature images and the classification of each projected point \citep{yang2018segmentation}. Similarly, \citeauthor{zhao2018classifying} first generated a group of multiscale contextual images regarding the height, intensity and roughness of each point in the point cloud and then designed a multiscale convolutional neural network to generate classification results of input point clouds \citep{zhao2018classifying}. Nevertheless, these methods involve generating 2D feature images from 3D point clouds, and their classification performance suffers from information loss during 3D to 2D transformation.

\subsubsection{Point-based methods}~\\
Pioneering work PointNet \cite{qi2017pointnet} started the trend of direct application of deep neural networks on the irregular point cloud. PointNet formulated the network architecture by a stack of multilayer perceptron (MLP) for hierarchical per-point feature learning, and a global pooling function was adopted to generate the global feature vector for object classification. Their experiments on various 3D point cloud recognition tasks demonstrate the great power of the PointNet model for point feature learning. \cite{qi2017pointnet++} further improved the performance with a multiscale point feature learning architecture by introducing a set abstract module and a feature propagation module for point set downsampling and upsampling. A hierarchical neural network was developed by recursively applying a unit PointNet on each grouped local region.

Considering the great success of PointNet \citep{qi2017pointnet} and PointNet++ \citep{qi2017pointnet++}, recent 3D point cloud classification and segmentation methods mostly build their methods upon PointNet-like architectures. For example, SO-Net~\cite{li2018so} explored a self-organizing map (SOM) to model the spatial distribution of point clouds and conducted hierarchical feature extraction on individual points and SOM nodes. PointSIFT~\cite{jiang2018pointsift} developed a new convolution module that encodes neighbor information under different orientations and scales using a SIFT-like operator. KC-Net~\cite{shen2018mining} introduced a kernel correlation-based convolution module to capture the local geometric structures of a point cloud. PointCNN \cite{li2018pointcnn} proposed a new general framework for point feature learning by transforming the input point sets into latent and potentially canonical order, thus enabling the leveraging of spatially local correlation.

For airborne LiDAR point clouds, \citeauthor{yousefhussien2018multi} proposed a 1D fully convolutional network that takes as input both raw coordinates of point clouds and three additional spectral features extracted from 2D georeferenced images for pointwise classification in an end-to-end fashion. Similarly, \citeauthor {wang2018deep} proposed a deep neural network with spatial pooling to achieve point cloud classification. Their method first extracted per-point features using a shared MLP network. Then, a spatial max-pooling layer was employed to aggregate the per-point features into the cluster-based features. Finally, another MLP network followed by a softmax layer classifier was used for point cloud classification \citep{wang2018deep}. \citeauthor{wen2019directionally} proposed a direction-constrained convolution module for point feature learning and designed a multiscale fully convolutional network for point cloud classification \citep{wen2019directionally}.

Although these point-based methods have achieved remarkable performance for point cloud classification, they cannot model the density variation in input point clouds. In this paper, we propose a novel density-aware density network (DANCE-NET) for end-to-end semantic classification of 3D point clouds. Compared to existing point-based methods, the proposed model can extract density-aware features and regularize the global semantic context with a context encoding module. We detail our method in the next section.

\section{Methods}\label{Methods}

In this section, we present the design for our end-to-end density-aware networks with context encoding (DANCE-NET) model. Our DANCE-NET model introduces a generalized convolution operation that can directly process point clouds without any form of projections. In Section 3.1, we introduce a density-aware novel convolution operation and compare it with the traditional image convolution. In Section 3.2, we describe how we incorporate kernel density estimation to rebalance learned weight functions in point convolution. In Section 3.3, we propose a novel auxiliary task for predicting contexts using global features to further regularize our model.

\subsection{Density-aware convolution}\label{sc_de}
Before going to our density-aware convolution module, we first revisit the general form of convolution operations. Generally, the convolution operation between two continuous functions $f \in \mathcal{R}^d$ and $g \in \mathcal{R}^d$ can be formulated as
\begin{equation}
(f*g)(x)=\int _{-\infty }^{\infty }f(\tau )g(x-\tau )\,d\tau     \label{eq:1}
\end{equation}

Commonly used 2D convolution can be regarded as a discrete version of Eq. \ref{eq:1}, where images can be interpreted as 2D discrete functions and the kernel functions are defined on gridded local windows. The 2D convolution between the input image $\mathcal{I}$ and kernel weights $\mathcal{K}$ can be formulated as
\begin{equation}
\mathcal{F}_{conv}(x, y) = \sum _{w=0}^{W}{\sum _{h=0}^{H}{\mathcal{K} (w,h)*\mathcal{I}(x-w,y-h)}}  \label{eq:2}
\end{equation}
where $x$ and $y$ denote the pixel location, $W$ and $H$ denote the size of the kernel window.

A point cloud of size $N$ can be represented by a set $S = \{p_i,...,p_n\}$, where each point $p_i = (x, y, z, f)$ contains the 3D coordinates and its features, such as color and surface normal. In contrast to image pixels that are located in fixed grids, the points of a 3D point cloud are scattered in $\mathbb{R}^3$, which can take an arbitrary continuous value. Therefore, there does not exist a fixed kernel pattern that can be applied to each point for feature extraction, which makes conventional 2D convolution inapplicable for an irregular 3D point set.

To generalize convolution operations to irregular 3D point sets, recent studies have proposed various convolution operations on 3D point clouds. The commonly used form can be expressed as
\begin{equation}
{FConv}(p_i) = \sum_{p_j \in R_i}\ \mathcal{K}(p_i, p_j)\mathcal{P}(p_i)
\end{equation}where for each $p_i \in S$, we find all its neighborhood points $p_j$ in the local region $R_i$ centered at $p_i$ with the k-nearest neighbor. $\mathcal{K}(p_i, p_j)$ measures the correlation between point $p_i$ and its neighbor point $p_j$. For example, KC-Net \citep{shen2018mining} used Gaussian kernel to account for function $\mathcal{K}(p_i, p_j)$. \citeauthor{wang2018deep} used MLP to implement the kernel function $\mathcal{K}(p_i, p_j)$, which enables learning of a parametric continuous convolution \cite{wang2018deep}.

In this paper, we focus on enhancing this convolution operation from the perspective of the spatial distribution of point clouds. Our key observation is that the irregular structure of 3D points gives rise to oversampling and undersampling of certain regions, resulting in disparate densities across the convolution kernel.

An analogy of this disparity in 2D images receives multiple values for one pixel. This creates a bias when we update the kernel, putting excessive weights on the oversampled regions and vice versa. In digitized images, pixels are placed evenly on a 2D grid where distances between neighboring pixels are constant; every cell in the grid has exactly one value, and there are no empty cells. This results in uniform density across the grid and, unfortunately, does not apply to irregular 3D point clouds.



\begin{figure}[t]
\centering
\includegraphics[width=8cm]{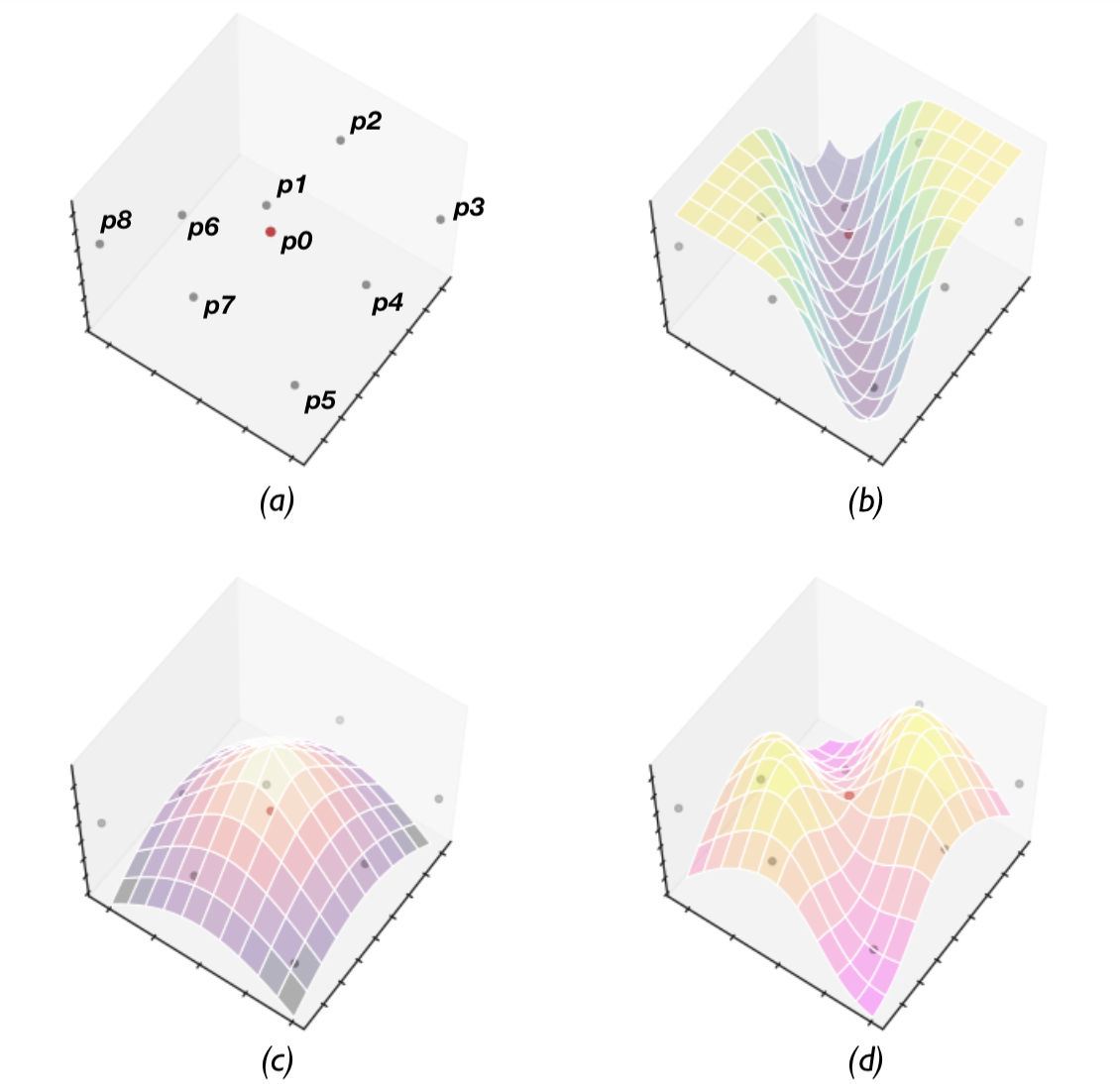}
\caption{Illustration of the proposed density-aware convolution module. (a) shows the sampled point set in the local region centered at $p_0$; our convolution module uses the relative distance between $p_0$ and other points for prediction (b) the continuous inverse density function. Then, we multiply the weight function (c) learned from the training data with the inverse density function (b) and produce (d) which approximates the weight function for the same scene but uniformly sampled.}
\label{fig_conv}
\end{figure}

To counter this imbalance, we introduce a novel convolution operation that applies the estimated densities in local regions to rescale the kernel functions. We formulate this new operation as follows:
\begin{equation}
{DConv}(p_i) = \sum_{p_j \in R_i} \mathcal{D}(p_i, p_j)\mathcal{K}(p_i, p_j)\mathcal{P}(p_i)
\end{equation}
where $\mathcal{D}(p_i, p_j)$ represents the density at point $p_j$, $\mathcal{D}(p_i, p_j)$ represents the kernel function for measuring the correlation between point $p_i$ and point $p_j$.

By formulating $\mathcal{D}$ and $\mathcal{K}$ as functions on point pairs $(p_i, p_j)$, we have a continuous domain in the region $R_i$, where we take point coordinates $(x, y, z) \in R_i$ as input, in contrast with the traditional image convolution, where only integer coordinates of pixels are accepted as input. The proposed convolution module can thus learn a network to approximate the continuous weights for convolution. Figure \ref{fig_conv} gives an illustration of the proposed density-aware convolution module.

In this paper, we use kernel density estimation (KDE) and compute an inverse density function $\mathcal{D}(p_i, p_j)$, followed by a nonlinear transformation implemented with MLP to map this density function to high-level embedding space. The nonlinear transformation here encourages the convolution module to adaptively determine whether to use the density estimates. The weights of the MLP here are shared across all the points to maintain the permutation invariance. For simplicity, we use another MLP to implement the kernel function $\mathcal{K}(p_i, p_j)$.


\subsection{Kernel Density Estimation (KDE)}\label{sc_density_estimation}
To compute the inverse density function $\mathcal{D}$, we think of the local region around any $p_i \in S$ as a ball with radius $\lambda$, and we apply the Parzen-Rosenblatt window method \citep{parzen1962} to fit density distributions along each of the axes inside the ball.
\begin{equation}
\mathcal{D}(p_i, p_j)_{KDE}={\frac {1}{Nh}}\sum _{p_j \in R_i}G{\Big (}{\frac {p_i-p_j}{h}}{\Big )}
\end{equation}
Here, we choose a multivariate Gaussian kernel $G$ with a smoothing factor $h$ and a normalization constant $N$, which is the number of points in the local region of $p_i$. We then pass these estimators through an MLP for nonlinear transformations, which provides a more refined approximation to the continuous density scale. The inverse density function is thus

\begin{equation}
\mathcal{D}(p_i, p_j) = \frac{1}{\mathcal{F}_{MLP}(\mathcal{D}(p_i, p_j)_{KDE})}
\end{equation}
Instead of relying on absolute distances, we estimate $\mathcal{D}$ using relative distances from the neighboring points $p_j \in \mathcal{N}_i$ to the centroid $p_i$, limiting the domain of the density function to $[-R, R]$, where $R$ is the radius of the local region. By fixing the domain, our model can generalize its density estimates to any arbitrary point with a local region of the same radius.

\begin{figure}[t]
\centering
\includegraphics[width=8cm]{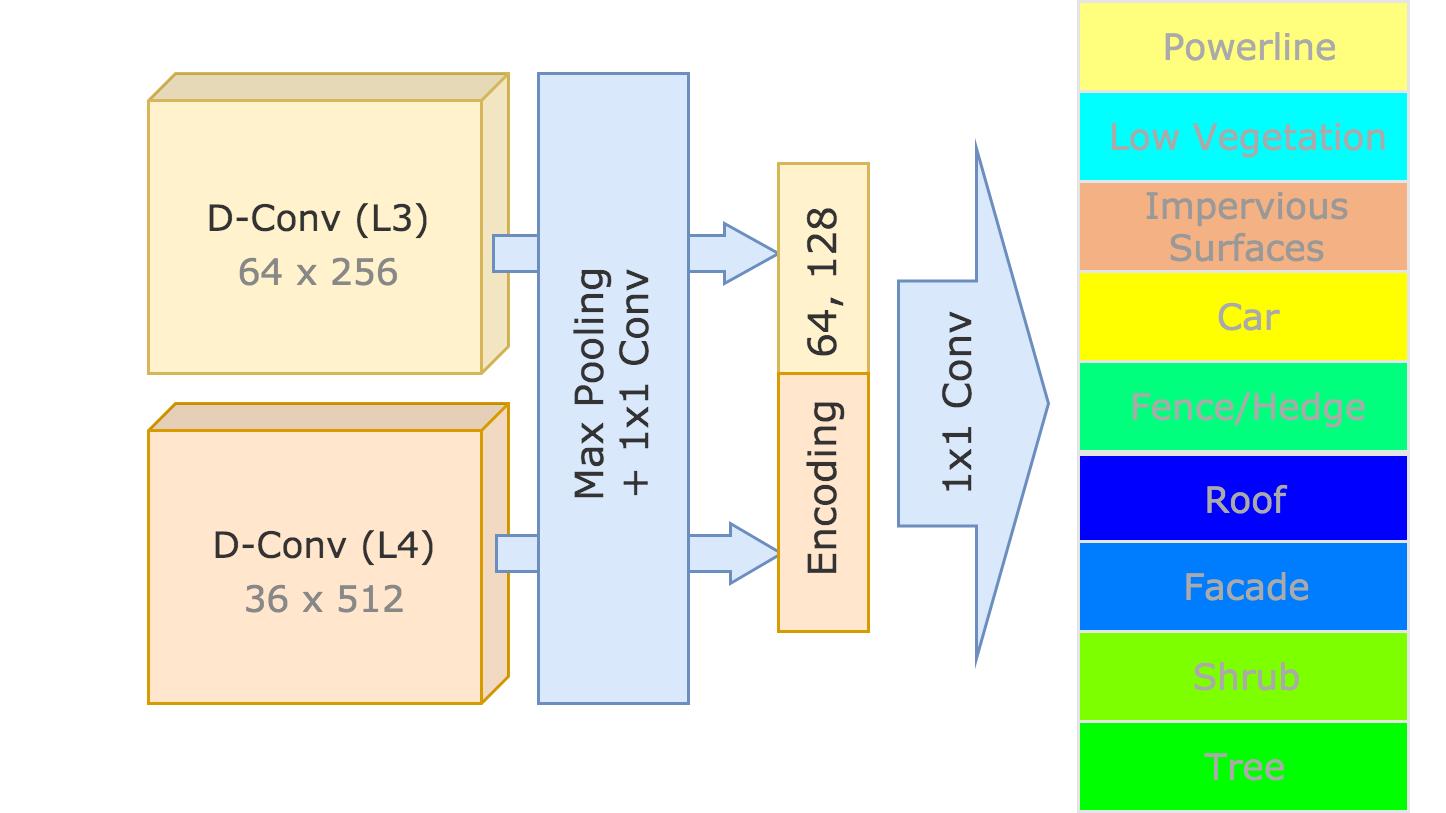}
\caption{Illustration of the proposed context encoding module.}
\label{fig_context_enc}
\end{figure}

\begin{figure*}[t]
\centering
\includegraphics[width=17cm]{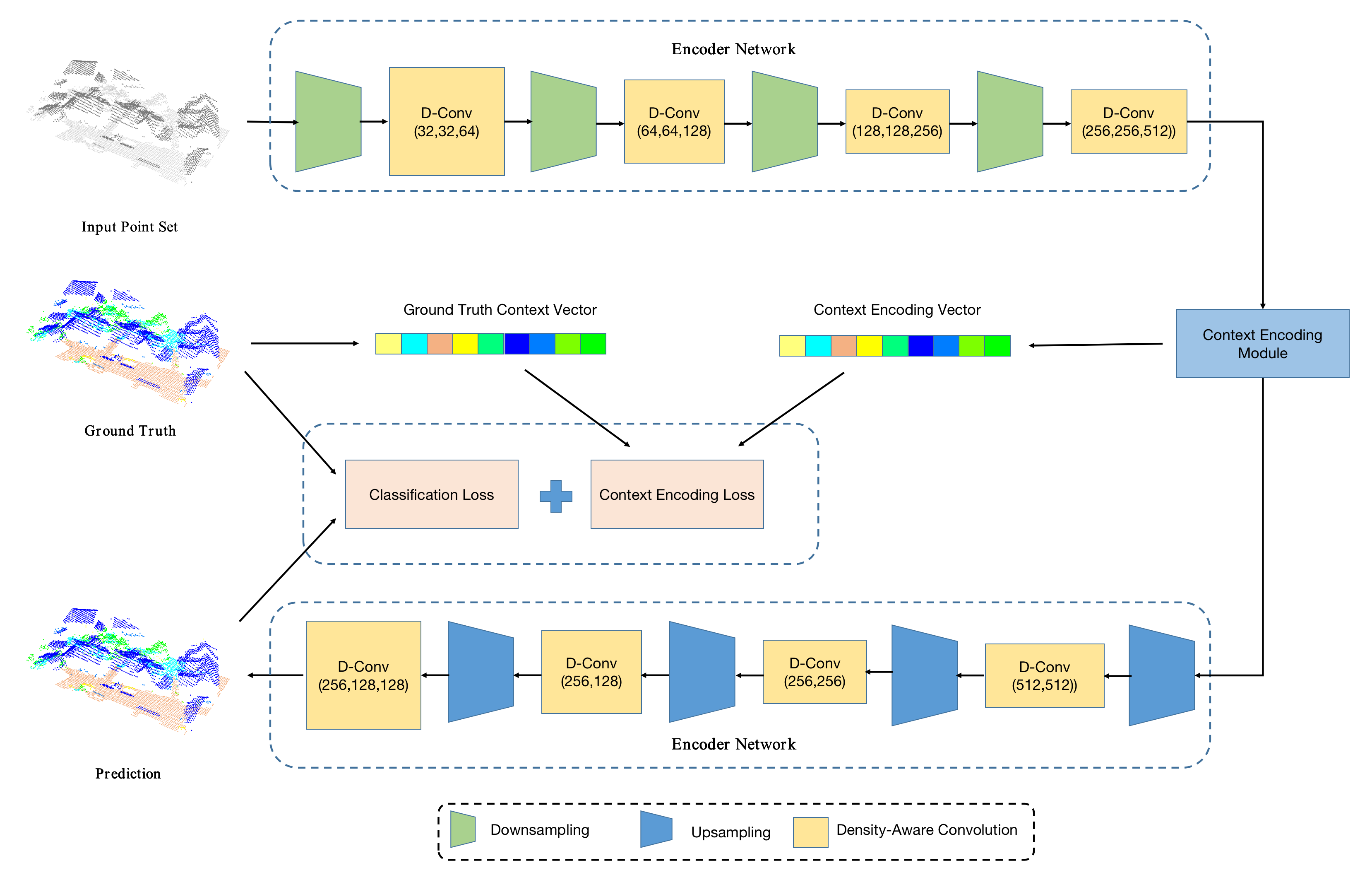}
\caption{Overview of the proposed method for airborne point cloud classification. Our model starts with an encoder network to extract high-level semantic features using a newly proposed density-aware convolution module. Then, a context encoding module was adopted to learn a global context encoding, which was further fed into a regularization module to enforce the predicted context encoding to be aligned with the ground truth encoding. Finally, a decoder network with a successive density-aware convolution module and upsampling block were used to generate per-point classification results.}
\label{fig_pipeline}
\end{figure*}

\subsection{Context Encoding Module}\label{sc_context_encoding}

A common problem in point cloud classification is that different object classes have very different numbers of training samples\citep{gustavo2004}. This could occur by nonuniform sampling where homogeneous clusters are oversampled or simply by the natural distribution of these classes. Regardless of the cause, this discrepancy in the number of training samples for each class introduces an issue where the classes with small numbers of samples are underfitted and are likely to be misclassified as the most frequent class in the process of maximizing accuracy. 

In this section, we propose an auxiliary task for predicting the context of the input point cloud to help guide the segmentation task. Given a patch of point set $S$ as a scene, we define the context vector as an indicator for the presence of different classes in this specific scene. Given a batch of points $B_i = \{p_1,...,p_N|\ p_j = (x, y, z, c) \in S\}$, where $c$ is the categorical label for each point, we define the ground truth global context vector $C^{gt}$ as in Eq.\eqref{eq:6}
\begin{equation}
    C_i^{gt} = 
    \begin{cases}
        1 & \text{if} \ \ \exists p_j \,\ s.t. \,\ c_j = i \\
        0 & \text{otherwise}
    \end{cases}      
    \label{eq:6}
\end{equation}
where $|C^{gt}| = k$ denotes the total number of classes, $C_i^{gt}$ is the $i$-th element of $C$. The resulting $C^{gt}$ represents the set of all possible contexts with cardinality $|S_{scene}| = 2^k$.

With this formulation of context, an underrepresented class, if present in the scene, will take the same value of 1 in the context vector $C$ as with other more frequently sampled classes. Therefore, predicting the context vector becomes a task of predicting the statistical property on the existence of different classes in the scene. This, in turn, regularize our classification network by giving larger weights to the undersampled classes and pushing the model to evaluate with a global view of the input set.

To generate the context vector prediction, we incorporate multiscale global feature learning where the network leverages features learned from multiple point densities and merges them using nonlinear transform into a compact encoding. This encoding is then fed to a multilayer perceptron (MLP) to predict a context vector $C$, which indicates the probability of the existence of each category. Figure \ref{fig_context_enc} illustrates the mechanism of our proposed context encoding module.

We use binary cross entropy loss between the predicted context vectors $C_i$ and the ground truth vectors $C_i^{gt}$ to regularize the context information, calculated as
\begin{equation}
\mathcal{L}_{ctx} = \sum_{i=1}^{|C|}C_i^{gt} * log(C_i)
\end{equation}
We then add this context regularization term to our final loss function and calculated as
\begin{equation}
\mathcal{L} = \mathcal{L}_{cls} +  \lambda *\mathcal{L}_{ctx} 
\end{equation}
\begin{equation}
\mathcal{L}_{cls} = \sum_{j=1}^{N}\sum_{c=1}^{|C|} [y_{jc} logp_{jc} + (1-y_{jc}) log(1-p_{jc}) ]
\label{eq_cls}
\end{equation}
where $\mathcal{L}_{cls}$ and $\mathcal{L}$ denote the classification loss and the final loss function, respectively. $\lambda$ denotes a hyperparameter to balance the classification loss and context encoding loss. We discuss the effect of different $\lambda$ in Section \ref{sc_hyper_param} and Section \ref{sc_effect_loss}.

\subsection{Network Architecture}\label{sc_net_arch}
The design of our DANCE-NET follows the widely used encoder-decoder architecture, where the input point clouds are first downsampled multiple times to learn hierarchical feature embeddings. Then, the learned embeddings are upsampled back to the original point sets for the semantic point classification. A context encoding module is integrated between the downsampling and upsampling path for the context encoding and regularization.

\begin{figure*}[t]
\centering
\includegraphics[width=17cm]{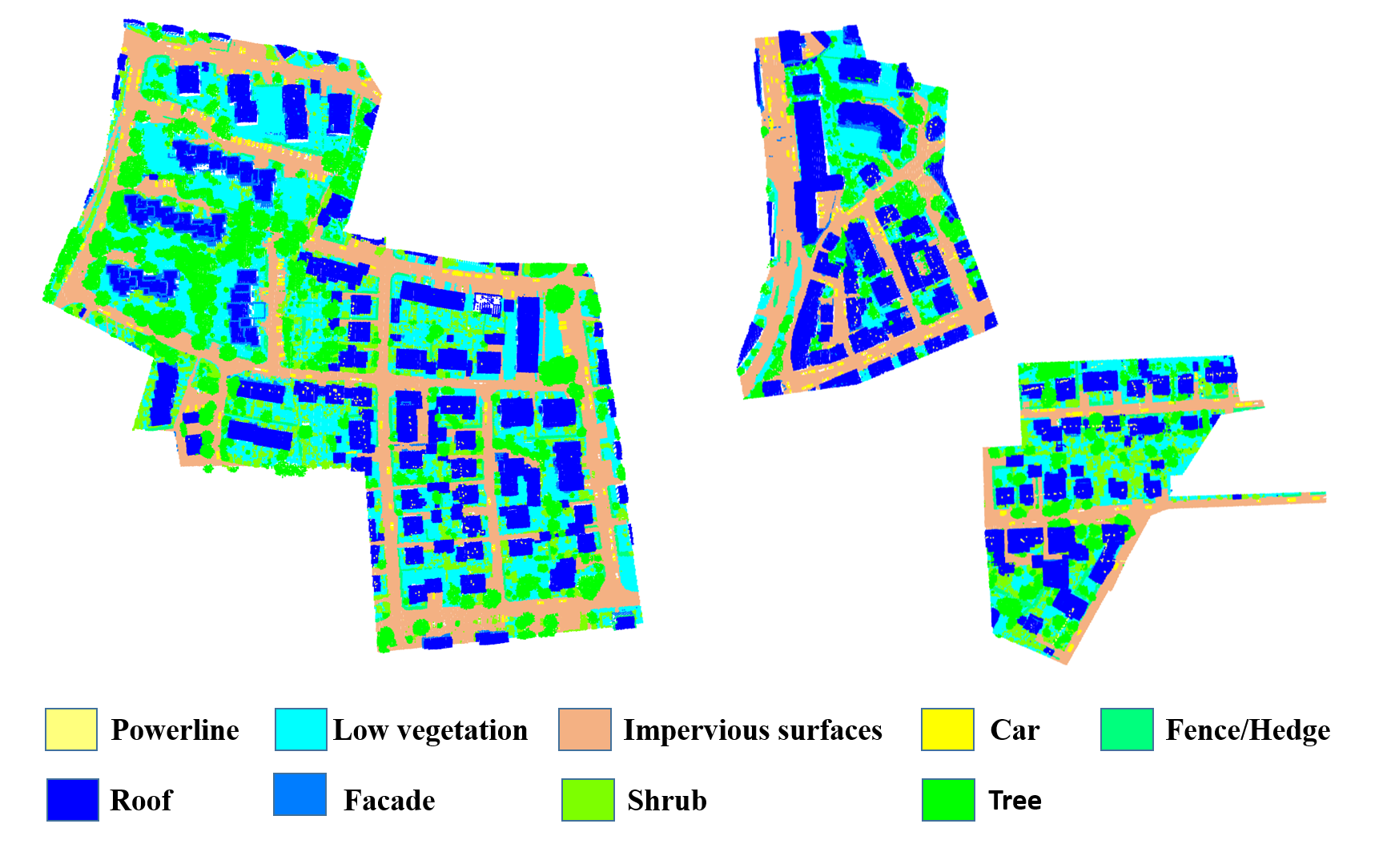}
\caption{
}
\label{fig_Dataset}
\end{figure*}

\subsubsection{Downsampling and upsampling blocks}~\\
Following the success of neural networks that learn high-level features from multiple scales, we build our model with downsampling blocks, producing point feature sets that are successively sparser and more complex, and upsampling blocks that interpolate the sparse set back to the original points that are introduced in PointNet++\citep{qi2017pointnet++}. We present a short explanation for the two types of blocks below.

The downsampling block takes as input a point set $S=\{p_1,...,p_n \\ | \ p_i = (x, y, z, f)\}$ where $x, y, z$ are coordinates of the point and $f$ is the feature vector, and outputs a new set $\hat{S} \subseteq S$ that contains a reduced number of points. In other words, the size of the output for the $i$-th downsampling block $|S_i|$ is less than or equal to $|S_{i-1}|$. To select the candidates for this new set $S_i$, we use the farthest point sampling, which selects the set of points that each of the selected points has the largest distance possible from the rest of the set. After selecting the candidates, we group the points in their respective local regions and feed them to our convolution module, obtaining density-weighted embeddings for each candidate. We define local regions with a ball query with a hyperparameter on the search radius, which we discuss further in Section \ref{sc_hyper_param}.

The upsampling block takes as input two point sets $S_i$ and $S_{i-1}$, which correspond to the output of the $i$-th and $(i-1)$-th downsampling blocks. It generates a point set $\bar{S} \supseteq S_i$ that contains the same points as in $S_{i-1}$ but has different features. In the upsampling process, we first take both sets and compute the distance between the known points $S_{i}$ and the points $S_{i-1}$ that we want to interpolate to. We normalize the inverse of these distances to the range of $(0, 1)$ and use it to scale the interpolated point features. Instead of directly predicting a continuous inverse density function for interpolation, we adopt this straight forward rescaling method to reduce the number of parameters in the model and still ensure that the embeddings are reweighted by some measure of density. Finally, we concatenate the feature vectors in $S_{i-1}$ with those of the interpolated set $\bar{S_i}$ for $1 \times 1$ convolution and extract high-level features of the desired dimension.

\subsubsection{Overall architecture}~\\
To present our network in more detail, the downsampling stage consists of four downsampling blocks which reduce the size of the input point set to 1024, 256, 64 and 36 sequentially, with each downsampling block having a channel sizes of 64, 128, 256 and 512, respectively. The point features outputted by the third and fourth blocks are fed into the context encoding module where each of the block's output is max-pooled and condensed by a 1x1 convolution to feature maps with 1/4 of the original channel size. We then concatenate these encodings, feed the result to an MLP and compute the binary cross entropy loss (BCE) with the ground truth context. In the upsampling stage, we use four upsampling blocks to increase the size of the point set to 64, 256, 1024 and the original size. To preserve low-level information, we incorporate the features learned in the downsampling stage and use skip connections to link them to upsampling blocks of the same output size. For example, the first upsampling block will first interpolate the 36-dimensional point features of the fourth downsampling block to 64-dimensional and then concatenate them with the 64-dimensional point features learned by the third downsampling block. This incorporation of low-level features has seen various successes in previous works for 2D segmentation, including performance improvement and faster convergence.
\citep{ronneberger2015u,badrinarayanan2017segnet}.

\section{Experiments and Results}\label{Experiments}
In this section, we conduct experiments to demonstrate the effectiveness of the proposed DANCE-NET model for airborne LiDAR point cloud labeling. We introduced the experimental dataset and data prepossessing in Section \ref{Experiment_Data} and Section \ref{sc_preprocessing}, respectively. Evaluation metrics were given in \ref{Methods_evaluation}. In Section \ref{sc_hyper_param}, we discussed the effect of different hyperparameter configurations. In Section \ref{Experiment_Setup}, we presented the experimental setups. The classification results of our model with optimal hyperparameters are given in Section \ref{Experiment_Result}.

\subsection{Experimental Dataset}\label{Experiment_Data}
We conducted our experiments on the commonly used International Society for Photogrammetry and Remote Sensing (ISPRS) 3D labeling dataset \citep{niemeyer2014contextual}. This dataset contains airborne LiDAR point clouds covering three isolated areas of Vaihingen city. These point clouds were captured by a Leica ALS50 system, with an average flying height of 500 meters above the ground and a field of view of 45 degrees \citep{cramer2010dgpf}. Each point in the dataset was labeled in 9 semantic categories, including powerline, low vegetation (low\_veg), impervious surface (imp\_surf), car, fence/hedge (fen/hed), roof, facade, shrub, and tree.

Following the standard settings of the ISPRS 3D labeling contest, the entire dataset was divided into two parts. The first scene (see Figure \ref{fig_Dataset} left) with 753,876 points was used as the training dataset, and the other two scenes (see Figure \ref{fig_Dataset} right) with 411,722 points were used as the test dataset. Detailed information about each scene is given in Table \ref{table_Dataset_number}. For each scene, a simple ASCII file with XYZ, reflectance, return count information, and point labels were provided in an ASCII file.

\begin{table}[h]
\begin{center}
\caption{Number of points in each object category of the training and test dataset.}
\label{table_Dataset_number}
\begin{tabular}{l c c}
\hline
Categories & Train & Test
\\
\hline
Powerline & 546 & 600\\
Low vegetation & 180,850 & 98,690\\
Impervious surfaces & 193,723 & 101,986\\
Car &  4,614 & 3,708\\
Fence/Hedge & 12,070  & 7,422\\
Roof & 152,045  & 109,048\\
Facade & 27,250 & 11,224\\
Shrub & 47,605  & 24,818\\
Tree & 135,173  & 54,226\\
\hline
Total & 753,876 & 411,722\\       
\hline
\end{tabular}
\end{center}
\end{table}

\subsection{Data prepossessing}\label{sc_preprocessing}
\begin{figure}[t]
\centering
\includegraphics[width=8cm]{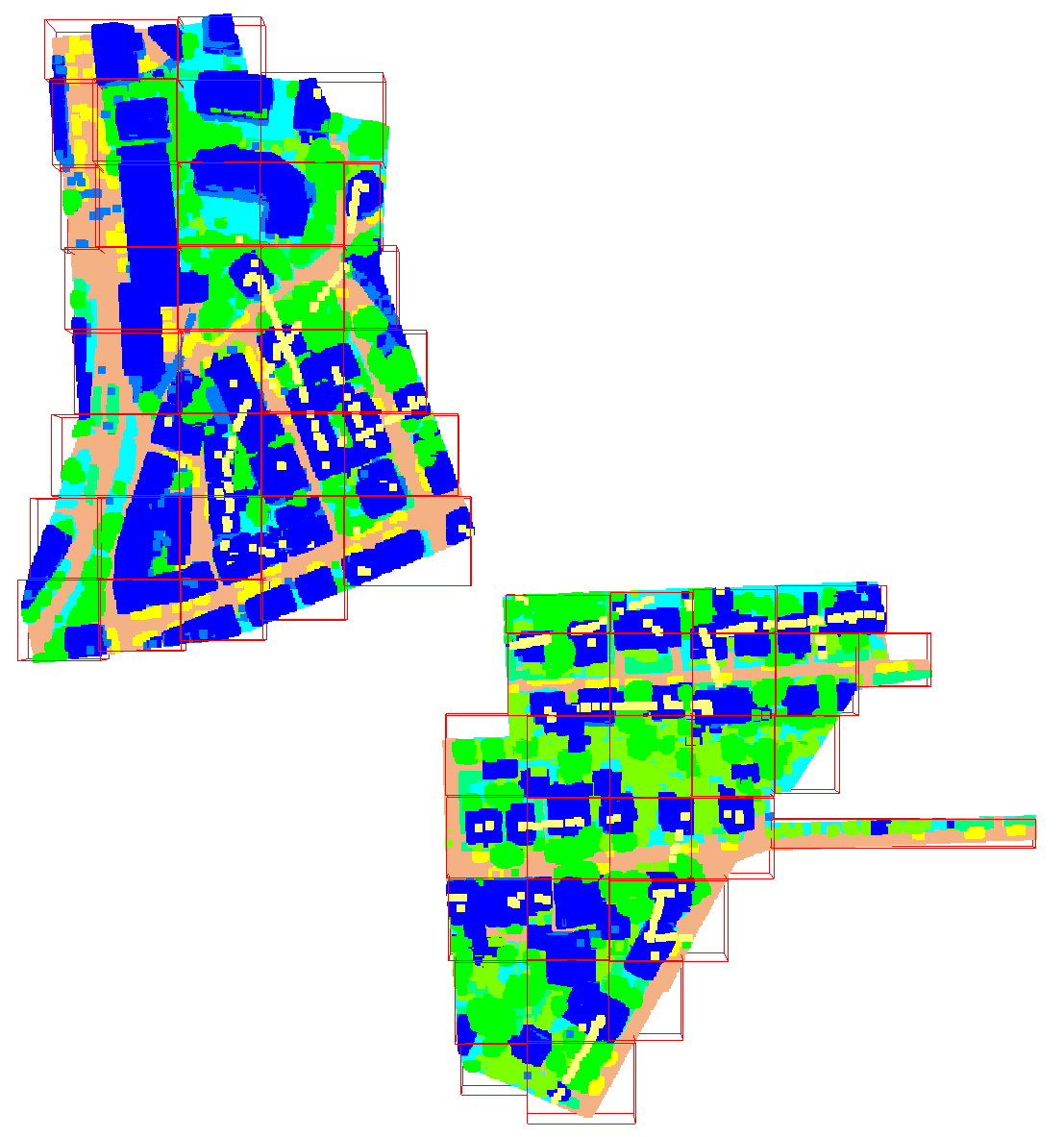}
\caption{
}
\label{fig_blocks}
\end{figure}

The original ISPRS 3D labeling dataset covers a large area with each scene containing more than 100 K points. Such a large point set cannot be directly fed into our model for network training due to the limited GPU memory. To facilitate training, we divided the training scene into small patches and only used a small batch of patches for model training at each training step (see Figure \ref{fig_blocks} for illustration). More specifically, we divided the training and test scenes into regular blocks of 30 m*30 m grids in the horizontal direction. Note that each block contains a different number of points. During training, we randomly select a group of blocks and sample a fixed number of points (e.g., 8192 points) from each block to formulate a batch of input point sets. To further improve the robustness of our model and reduce the risk of overfitting, we randomly drop a certain number of points at each block during the training stage. By default, we set the dropout ratio to 12.5\% in our experiments. It is worth mentioning that, due to the irregular boundary of each scene, some edge blocks may contain a small number of points and thus cannot be effectively classified due to limited geometric information. We thus merge those edge blocks with their surrounding blocks with more points.

In the inference stage, all points in each test block were directly fed into our model for point labeling. Here, we do not need point sampling for each block because our model by nature is a fully convolutional network and can receive an arbitrary size of input point sets. We can easily merge the label predictions from each test block to generate the final prediction results.

Further investigating the ISPRS 3D labeling dataset, one can find out that there is quite a different number of points for each object category. For example, as shown in Table \ref{table_Dataset_number}, there are only 546 powerline points in the whole training scene, while the number of impervious surface points amounts to 193,723. Directly training on this unbalanced dataset may cause the issue where the classes with small numbers of points get under-fitted and are likely to be misclassified by deep neural networks. To address this issue, we introduce the context encoding module in Section \ref{sc_context_encoding} to force context prediction to align with the ground truth. Here, we introduce a simpler strategy to achieve the same goal. With the intention to enforce the model to pay more attention to those categories with a small number of points, we added a category-specific weight coefficient for each category to the loss function of our DANCE-NET model. The balance weight for each category is calculated by the inverse of the logarithm of the ratio of each category, formulated as Equation \ref{layer_function}.

\begin{equation}
W_{i} = \frac{1}{\ln (\alpha + \frac{{N}_{i}}{{\sum_{i=1}^{|C|} {N}_{i}}})}
\label{layer_function}
\end{equation}

where $W_i$ denotes the weight of the $i$th category, $N_i$ represents the number of points in the $c$th category, $|C|$ denotes the total number of categories, $\alpha$ denotes the coefficient for class balance. We set $\alpha$ to 1.2 according to experimental evaluation. By integrating the category-specific weights, we reformulate Eq. \ref{eq_cls} as:
\begin{equation}
\mathcal{L}_{cls} = \sum_{j=1}^{N} w_j * \sum_{c=1}^{|C|} [y_{jc} logp_{jc} + (1-y_{jc}) log(1-p_{jc}) ]
\label{loss_function}
\end{equation}
where $N$ denotes the number of sampling points in each training block, $y_{jc}$ and $p_{jc}$ denote the ground truth label and predicted probability of the $j$th point on the $c$th category, and $w_j$ denotes the balance weight for the $j$th sampling point and is calculated as $w_j:= W_k | y_{jk}=1$.

\subsection{Evaluation Metrics}\label{Methods_evaluation}
We evaluate the performance of our model for airborne LiDAR point cloud classification using two metrics, i.e., overall accuracy (OA) and F1 score. In general, overall accuracy measures the classification accuracy for all categories as a whole, which is defined as the ratio of corrected classified points in the total test point sets. While F1 score deals with each category separately and considers both the precision and recall values. The F1 score is generally more suitable for evaluating unevenly distributed datasets. The calculation of overall accuracy, precision, recall and F1 score are defined as follows:
\begin{equation}
OA = \frac{tp+tn}{tp+tn+fp+fn}
\end{equation}
\begin{equation}
precision = \frac{tp}{tp+fp}
\end{equation}
\begin{equation}
recall = \frac{tp}{tp+fn}
\end{equation}
\begin{equation}
F1 = 2*\frac{precision*recall}{precision+recall}
\end{equation}

where tp (true positive)/tn (true negative) denotes the number of positive/negative tuples that were correctly labeled by the classifier and fp (false positive)/fn (false negative) denote the negative/positive tuples that were incorrectly labeled by the classifier.

\subsection{Experimental Setup}\label{Experiment_Setup}

Our DANCE-NET method was implemented using the TensorFlow framework. During network training, we set the batch size to 6. The Adam optimizer with an initial learning rate of 0.01 was used to train our network, and we divided the learning rate by 2 every 3,000 steps. It took approximately 10 hours to train our model for 1,000 epochs until convergence on a TESLA K80 GPU. Our code will be released at \url{https://github.com/lixiang-ucas/DANCE-NET}.

In the downsampling and upsampling blocks, we set a base neighborhood size of 2 m for the input point cloud. In the downsampling stage, with the decrease of the size of the point cloud, we multiply the neighborhood size by 2 after each downsampling block. Similarly, in the upsampling stage, with the increase of the size of the point cloud, we divided the neighborhood size by two after each upsampling block. The bandwidth for KDE follows the same rules but has its base value set as 1 m.

\subsection{Hyperparameters Selection}\label{sc_hyper_param}
In this section, we conduct extensive experiments to determine the best hyperparameter configurations for our DANCE-NET model. The overall accuracy and average F1 score were used to evaluate the performance of different model configurations. We listed the performance of different model configurations in Table \ref{ta_params}.

As shown in Table \ref{ta_params}, our model achieved quite stable performance with different settings for the hyperparameters, including searching radius $R$ and $\lambda$. Moreover, our model achieved the best performance of 0.839 and 0.712 for overall accuracy and average F1 score when $R$ and $\lambda$ were set to 2 m and 1, respectively. In the following sections, we used this configuration as our default setting.

\begin{table}[h]
    \centering
    \begin{tabular}{c|c|c|c}
        \hline
        Search radius (R) & $\lambda$ & OA & Average F1 \\
        \hline
        1 & 0.8 & 0.829 & 0.707 \\
        2 & 0.8 & 0.836 & 0.705 \\
        4 & 0.8 & 0.834 & 0.702 \\
        \hline
        1 & 1 & 0.830 & 0.703 \\
        2 & 1 & \textbf{0.839} & \textbf{0.712} \\
        4 & 1 & 0.832 & 0.705 \\
        \hline
        1 & 1.2 & 0.830 & 0.708 \\
        2 & 1.2 & 0.830 & 0.705 \\
        4 & 1.2 & 0.830 & 0.708 \\
        \hline
    \end{tabular}
    \caption{Classification performance with different hyperparameter configurations.}
    \label{ta_params}
\end{table}

\begin{figure}[h]
\centering
\includegraphics[width=8cm]{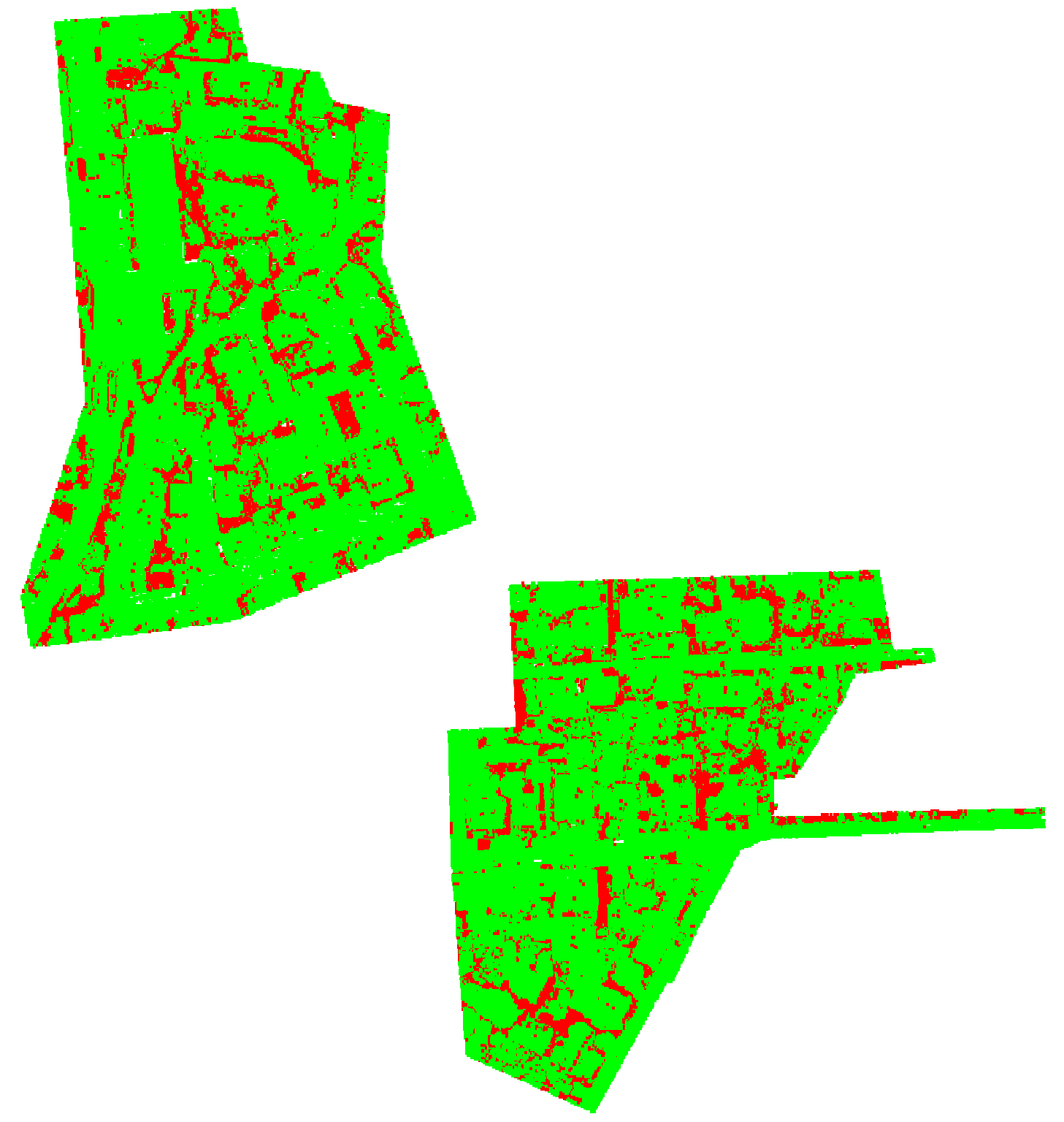}
\caption{
}
\label{fig_Testset_split}
\end{figure}

\subsection{Classification Results}\label{Experiment_Result}
\begin{figure*}[!b]
\centering
\includegraphics[width=17cm]{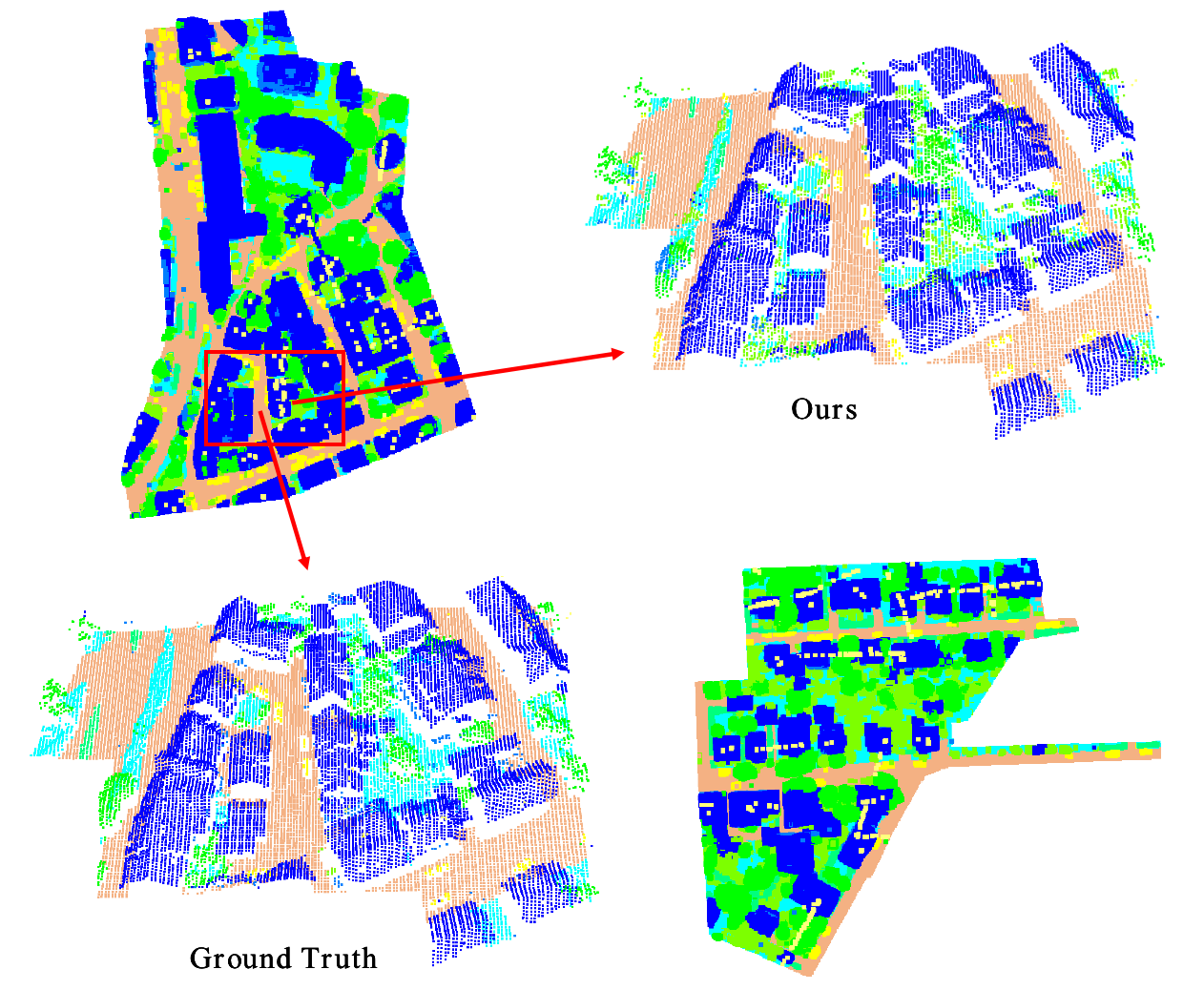}
\caption{
}
\label{fig_Result_overview}
\end{figure*}

After determining the best hyperparameters for the model, we employed Scene I of the ISPRS 3D Semantic Labeling Dataset to train the proposed DANCE-NET model until convergence. Then, each test block from Scene II and Scene III mentioned in Section \ref{sc_preprocessing} was directly fed into the model without sampling or dropout. The final point cloud classification results and error maps are shown in Figure \ref{fig_Testset_split} and Figure \ref{fig_Result_overview}. As shown in Figure \ref{fig_Testset_split} and Figure \ref{fig_Result_overview}, the proposed DANCE-NET model successfully generates the correct label predictions for most of the points in the test scenes.

\begin{table*}[!t]
\begin{center}
\caption{The classification confusion matrix of our DANCE-NET model. Precision, recall, and F1 score are also reported. Our model achieved an overall accuracy of 83.9\% and an average F1 score of 71.2\%. }
\label{table_confusion_matrix}
\begin{tabular}{l c c c c c c c c c}
\hline
Categories & powerline & low\_veg & imp\_surf & car & fence\_hedge & roof & facade & shrub & tree 
\\
\hline
powerline & \textbf{0.668} & 0.000 & 0.000 & 0.000 & 0.000 & 0.001 & 0.001 & 0.000 & 0.000 \\
low\_veg & 0.002 & \textbf{0.771} & 0.051 & 0.042 & 0.075 & 0.014 & 0.053 & 0.121 & 0.015 \\
imp\_surf & 0.000 & 0.091 & \textbf{0.946} & 0.008 & 0.023 & 0.001 & 0.009 & 0.006 & 0.002 \\
car & 0.000 & 0.001 & 0.001 & \textbf{0.745} & 0.02  & 0.000 & 0.007 & 0.009 & 0.001 \\
fence\_hedge  & 0.000 & 0.006 & 0.000 & 0.013 &\textbf{ 0.283} & 0.000 & 0.003 & 0.022 & 0.002 \\
roof & 0.228 & 0.009 & 0.001 & 0.026 & 0.047 & \textbf{0.937} & 0.178 & 0.055 & 0.026 \\
facade & 0.015 & 0.003 & 0.000 & 0.010 & 0.017 & 0.010 & \textbf{0.535} & 0.019 & 0.011 \\
shrub & 0.008 & 0.095 & 0.001 & 0.156 & 0.399 & 0.011 & 0.148 & \textbf{0.580} & 0.109 \\
tree & 0.078 & 0.024 & 0.000 & 0.000 & 0.136 & 0.026 & 0.067 & 0.188 & \textbf{0.834 }\\
\hline
Precision & 0.668 &  0.771 &  0.946 &  0.745 &  0.283 &  0.937 &  0.535 &  0.580 &  0.834 \\ 
Recall & 0.700 &  0.866 &  0.910 &  0.802 &  0.606 &  0.942 &  0.690 &  0.398 &  0.795 \\ 
F1 score & 0.684 &  0.816 &  0.928 &  0.772 &  0.386 &  0.939 &  0.602 &  0.472 &  0.814  \\ 
\hline
\end{tabular}
\end{center}
\end{table*}

To quantitatively evaluate the classification performance, we calculated the classification confusion matrix, precision, recall and F1 score of each category and listed the results in Table \ref{table_confusion_matrix}. As we can see in this Table, our proposed model obtained F1 scores higher than 70\% for six of the categories, including powerline, low vegetation, impervious surfaces, car, roof, and tree.
In addition, our model achieved a reasonable classification performance on the facade category, while the classification performances on the fence/hedge and shrub categories were relatively lower.
As Table \ref{table_confusion_matrix} shows, most of the fence/hedge points were incorrectly classified as shrubs. One of the main reasons is that the fence/hedge category contains fewer points and presents similar spatial distribution and topological characteristics with the shrub category, which causes the model to not be fully trained and thus hinders the model from differentiating these two categories. Although powerline and car categories also have fewer points in the dataset, they present completely different characteristics from other categories and thereby acquired higher classification performance.
\section{Discussion}\label{sc_discussion}
In this section, we first demonstrated the effectiveness of the proposed density-aware point convolution module and context encoding loss module in sections \ref{sc_effect_density} and \ref{sc_effect_loss}. Then, in section \ref{Experiment_Comparison}, we compared the performance of our model with other state-of-art methods for Airborne LiDAR point cloud classification.

\subsection{Effect of Density Encoding}\label{sc_effect_density}
\begin{figure*}[t]
\centering
\includegraphics[width=17cm]{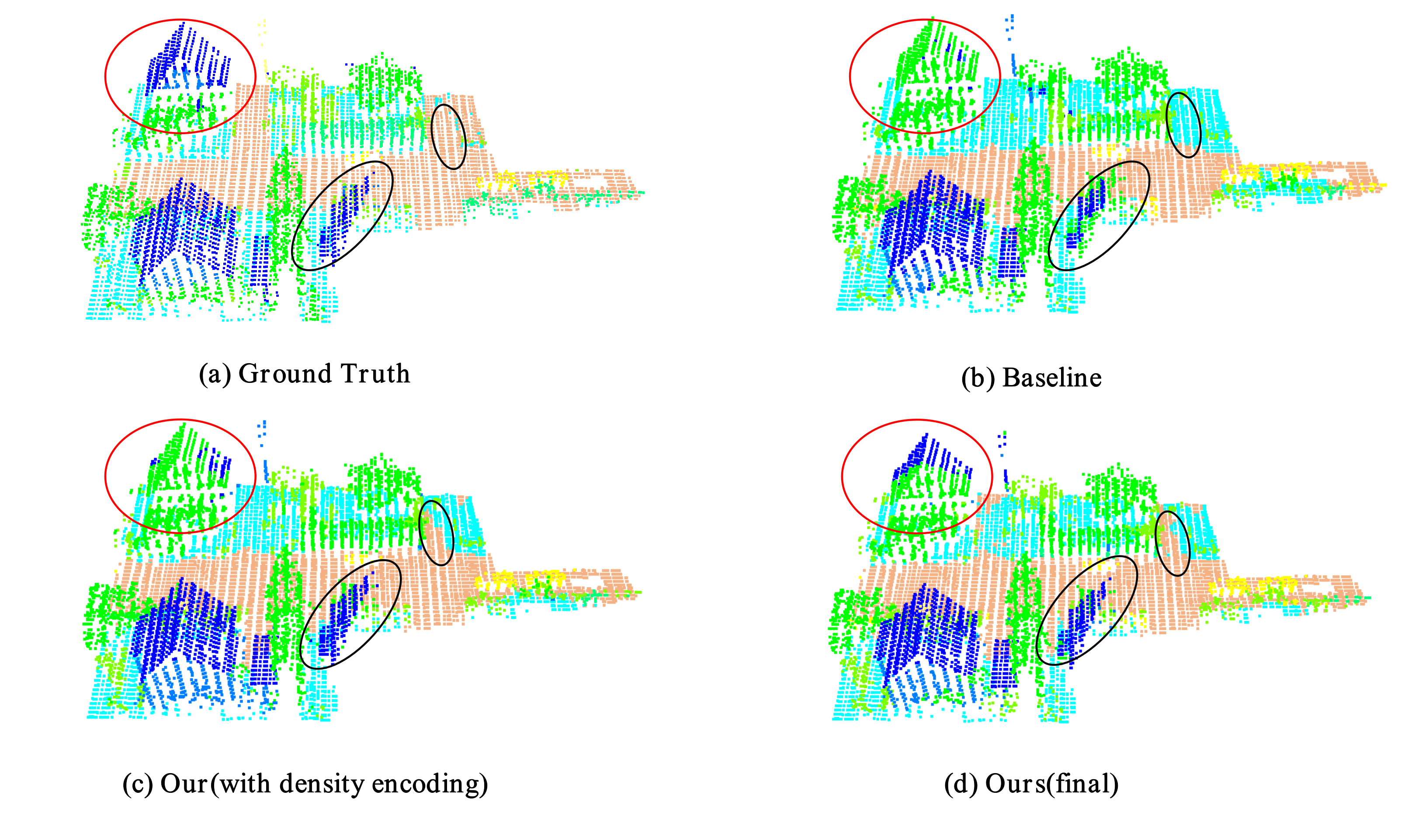}
\caption{The classification results with different model configurations. (a) Ground truth labels, (b) baseline model, (c) baseline model with the proposed density-aware encoding convolution module and (d) our model with both density-aware encoding convolution module and context encoding module. The black circled parts highlight the difference between (b) and (c), and the red circled part highlights the differences between (c) and (d).}
\label{fig_ablation}
\end{figure*}

To demonstrate the effectiveness of the proposed density-aware point convolution module and context encoding module, we developed three models: (a) the model without the density encoding module and context encoding module, we marked this model as 'baseline', (b) the model with density encoding only, marked as 'ours (with density encoding)', (c) the model with both density encoding and context encoding, marked as 'ours (final)'. We listed the classification results of these models in Table \ref{tab_ablation}. Qualitative comparison results are shown in Figure \ref{fig_ablation}.

As shown in Table \ref{tab_ablation}, with the density-aware encoding convolution module, the proposed method improved the overall accuracy and average F1 score by 0.8\% and 2.1\%, respectively. Comparing Figure \ref{fig_ablation} (b) and Figure \ref{fig_ablation}(c), we can see that our model successfully corrected some misclassified roof points by introducing the density-aware encoding convolution module.

\begin{table}[h]
    \centering
    \begin{tabular}{c c c}
        \hline
        Method & OA & Average F1 \\
        \hline
        baseline & 0.825 & 0.681\\
        Ours (with density) & 0.833 & 0.702\\
        Ours (final) & 0.839 & 0.712\\
        \hline
    \end{tabular}
    \caption{Ablation analysis of our DANCE-NET model.}
    \label{tab_ablation}
\end{table}

\subsection{Effect of Context Encoding Module}\label{sc_effect_loss}
As shown in Table \ref{tab_ablation}, with the proposed context encoding module, our model obtained an additional improvement of 0.6\% and 1.0\% on overall accuracy and average F1 score, respectively. Comparing Figure \ref{fig_ablation} (c) and Figure \ref{fig_ablation}(d), we find that our model successfully corrected a large number of misclassified roof points and several impervious surface points.

\begin{table*}[h]
\begin{center}
\caption{Quantitative comparisons between our method and other state-of-art models on the ISPRS benchmark dataset. The numbers in the first nine columns of the table indicate the F1 scores for each category, and the last two columns indicate the overall accuracy (OA) and average F1 score (Average F1). The boldface text indicates the model with the best performance.}
\label{table_avgF1_comparion}
\begin{tabular}{l c c c c c c c c c c c}
\hline
Categories & power & low\_veg & imp\_surf & car & fence\_hedge & roof & facade & shrub & tree & OA & Average F1\\
\hline
UM & 0.461 & 0.790 & 0.891 & 0.477 & 0.052 & 0.920 & 0.527 & 0.409 & 0.779 & 0.808 & 0.590 \\
WhuY2 & 0.319 & 0.800 & 0.889 & 0.408 & 0.245 & 0.931 & 0.494 & 0.411 & 0.773 & 0.810 & 0.586 \\ 
WhuY3 & 0.371 & 0.814 & 0.901 & 0.634 & 0.239 & 0.934 & 0.475 & 0.399 & 0.780 & 0.823 & 0.616 \\
LUH & 0.596 & 0.775 & 0.911 & 0.731 & 0.340 & 0.942 & 0.563 & 0.466 & \textbf{0.831} & 0.816 & 0.684 \\
BIJ\_W & 0.138 & 0.785 & 0.905 & 0.564 & 0.363 & 0.922 & 0.532 & 0.433 & 0.784 & 0.815 & 0.603 \\
RIT\_1 & 0.375 & 0.779 & 0.915 & 0.734 & 0.180 & 0.940 & 0.493 & 0.459 & 0.825 & 0.816 & 0.633 \\
NANJ2 & 0.620 & \textbf{0.888} & 0.912 & 0.667 & 0.407 & 0.936 & 0.426 & \textbf{0.559} & 0.826 & \textbf{0.852} & 0.693 \\
WhuY4 & 0.425 & 0.827 & 0.914 & 0.747 & \textbf{0.537} & \textbf{0.943} & 0.531 & 0.479 & 0.828 & 0.849 & 0.692 \\
\hline
Ours  &  \textbf{0.684}  &  0.816  & \textbf{0.928}   & \textbf{0.772}  &  0.386 &  0.939  &  \textbf{0.602}  &  0.472  &  0.814 & 0.839 & \textbf{0.712} \\
\hline
\end{tabular}
\end{center}
\end{table*}

\begin{figure*}[h]
\centering
\includegraphics[width=17cm]{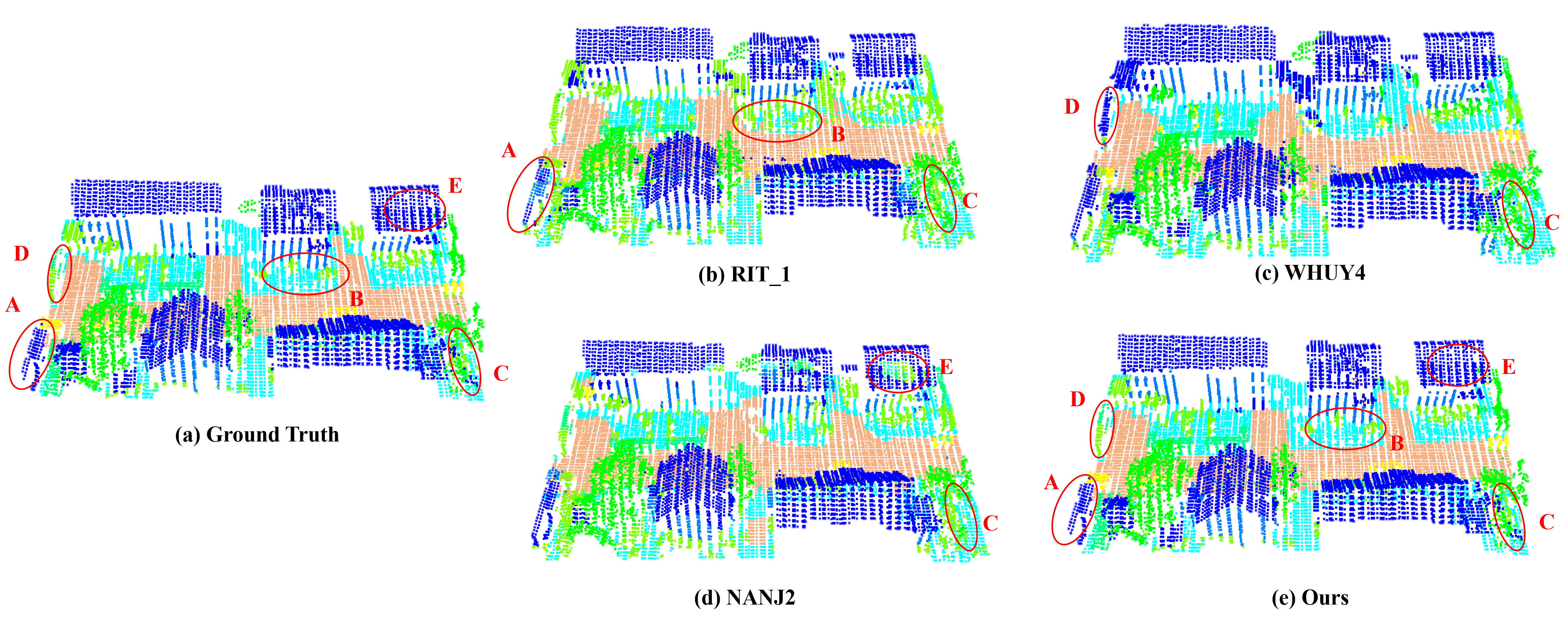}
\caption{Comparison of the classification results generated by RIT\_1, WhuY4, NANJ2 and our proposed DANCE-NET model in a selected complicated scene area. The red circle indicates areas where our model performs better than the comparing methods. Best viewed in color.}
\label{fig_Testset_compare}
\end{figure*}

\subsection{Comparison with other methods}\label{Experiment_Comparison}
To demonstrate the superiority of the proposed model, we compared it with other state-of-the-art methods that also report their performance on the ISPRS 3D labeling benchmark. The top eight models, i.e., UM \citep{horvat2016context}, WhuY2, WhuY3 \citep{yang2017convolutional}, LUH \citep{niemeyer2016hierarchical}, BIJ\_W \citep{wang2018deep}, RIT\_1 \citep{yousefhussien2018multi}, NANJ2 \citep{zhao2018classifying} and WhuY4 \citep{yang2018segmentation}, that reported the best performance on this benchmark were selected for performance comparison. Table  \ref{table_avgF1_comparion} lists the overall accuracy and F1 scores of our model and these comparing methods. As shown in Table \ref{table_avgF1_comparion}, the proposed model achieved a new state-of-the-art performance of 71.2\% on average F1 score, which surpassed the state-of-the-art WhuY4 \cite{yang2018segmentation} model by 2.0\%.
Additionally, our model achieved an overall accuracy of 83.9\%, which was comparable to the state-of-the-art NANJ2 \cite{zhao2018classifying} model with an overall accuracy of 85.2\%. Note that the NANJ2 method uses additional corresponding RGB features and hand-crafted features (e.g., intensity, roughness) as input for point cloud classification, while our model only uses raw XYZ coordinates as inputs. More specifically, the proposed model achieved remarkably higher performance in the powerline, impervious surface, car and facade categories.

We further evaluated the performance of our model by comparing the qualitative results with state-of-the-art methods. The top three methods on the ISPRS 3D labeling benchmark, i.e., RIT\_1 \citep{yousefhussien2018multi}, NANJ2 \citep{zhao2018classifying} and WhuY4  \citep{yang2018segmentation}, were selected for comparison. Figure \ref{fig_Testset_compare} shows the classification results generated by all comparing methods, and the red circle indicates where the comparing methods obtain incorrect predictions, and our model obtains correct predictions. For example, by comparing Figure \ref{fig_Testset_compare}(b), (c), (d) and (e), one can find that RIT\_1 misclassified some low vegetation points as tree and WHUY4 method misclassified some tree points as roof (area 'D'), while the NANJ2 method misclassified some roof points as tree (area 'E'). Our method obtained correct predictions for all challenging areas (including 'A', 'B', 'C', 'D', 'E').

One may also note that our model fell behind the state-of-the-art method on the overall accuracy. This is mainly because our model used a weighted loss to rebalance the learning process for each category, see Section \ref{sc_preprocessing}. Another reason is that we also use the average F1 score as the main indicator when determining the model hyperparameters rather than referring to overall accuracy. Considering that the ISPRS 3D labeling dataset contains a very different number of points for each category, focusing on overall accuracy may lead to those minority categories being ignored; we, therefore, focused primarily on the F1 score in this study.

\section{Conclusions}\label{sc_conclusion}
In this paper, we proposed a novel density encoding-aware network with a context encoding module for the task of 3D point cloud classification. To better model the spatial distribution of the 3D point set, a density-aware convolution module was designed by reweighting the learnable weights of convolution kernels using the pointwise density of each point. The proposed convolution model is thus able to fully approximate the 3D continuous convolution on unevenly distributed 3D point clouds. Based on the proposed convolution module, we further developed a multiscale fully convolutional neural network to achieve point cloud classification with hierarchical downsampling and upsampling blocks. Moreover, to address the issue of imbalanced class distribution of outdoor point clouds, we developed a context encoding module to force the predicted context to align with the ground truth context information. Experimental results on the ISPRS 3d labeling benchmark demonstrated the effectiveness of our proposed density-aware convolution module and context encoding module. The proposed model achieved new state-of-the-art performance on average F1 score by only taking the raw 3D coordinates and intensity as inputs.

\section*{ACKNOWLEDGMENTS}\label{ACKNOWLEDGEMENTS}
We would like to gratefully acknowledge the ISPRS for providing airborne LiDAR data.

{
\begin{spacing}{0.9}
\bibliography{bibliography} 

\begin{thebibliography}{xx}

\bibitem[Andersen et al., 2005]{andersen2005estimating}
Andersen, H.-E., McGaughey, R.~J. and Reutebuch, S.~E., 2005.
\newblock Estimating forest canopy fuel parameters using lidar data.
\newblock Remote sensing of Environment 94(4), pp.~441--449.

\bibitem[Axelsson, 2000]{axelsson2000generation}
Axelsson, P., 2000.
\newblock Dem generation from laser scanner data using adaptive tin models.
\newblock International archives of photogrammetry and remote sensing 33(4),
  pp.~110--117.

\bibitem[Babahajiani et al., 2017]{babahajiani2017urban}
Babahajiani, P., Fan, L., K{\"a}m{\"a}r{\"a}inen, J.-K. and Gabbouj, M., 2017.
\newblock Urban 3d segmentation and modelling from street view images and lidar
  point clouds.
\newblock Machine Vision and Applications 28(7), pp.~679--694.

\bibitem[Badrinarayanan et al., 2017]{badrinarayanan2017segnet}
Badrinarayanan, V., Kendall, A. and Cipolla, R., 2017.
\newblock Segnet: A deep convolutional encoder-decoder architecture for image
  segmentation.
\newblock IEEE transactions on pattern analysis and machine intelligence
  39(12), pp.~2481--2495.

\bibitem[Batista et al., 2004]{gustavo2004}
Batista, G. E. A. P.~A., Prati, R.~C. and Monard, M.~C., 2004.
\newblock A study of the behavior of several methods for balancing machine
  learning training data.
\newblock ACM SIGKDD Explorations Newsletter 6(1), pp.~20.

\bibitem[Chehata et al., 2009]{chehata2009airborne}
Chehata, N., Guo, L. and Mallet, C., 2009.
\newblock Airborne lidar feature selection for urban classification using
  random forests.
\newblock International Archives of Photogrammetry, Remote Sensing and Spatial
  Information Sciences 38(Part 3), pp.~W8.

\bibitem[Cheng et al., 2016]{cheng2016learning}
Cheng, G., Zhou, P. and Han, J., 2016.
\newblock Learning rotation-invariant convolutional neural networks for object
  detection in vhr optical remote sensing images.
\newblock IEEE Transactions on Geoscience and Remote Sensing 54(12),
  pp.~7405--7415.

\bibitem[Cramer, 2010]{cramer2010dgpf}
Cramer, M., 2010.
\newblock The dgpf-test on digital airborne camera evaluation--overview and
  test design.
\newblock Photogrammetrie-Fernerkundung-Geoinformation 2010(2), pp.~73--82.

\bibitem[Ene et al., 2017]{ene2017large}
Ene, L.~T., N{\ae}sset, E., Gobakken, T., Bollands{\aa}s, O.~M., Mauya, E.~W.
  and Zahabu, E., 2017.
\newblock Large-scale estimation of change in aboveground biomass in miombo
  woodlands using airborne laser scanning and national forest inventory data.
\newblock Remote Sensing of Environment 188, pp.~106--117.

\bibitem[Garc{\'\i}a-Guti{\'e}rrez et al., 2015]{garcia2015evolutionary}
Garc{\'\i}a-Guti{\'e}rrez, J., Mateos-Garc{\'\i}a, D., Garcia, M. and
  Riquelme-Santos, J.~C., 2015.
\newblock An evolutionary-weighted majority voting and support vector machines
  applied to contextual classification of lidar and imagery data fusion.
\newblock Neurocomputing 163, pp.~17--24.

\bibitem[Horvat et al., 2016]{horvat2016context}
Horvat, D., {\v{Z}}alik, B. and Mongus, D., 2016.
\newblock Context-dependent detection of non-linearly distributed points for
  vegetation classification in airborne lidar.
\newblock ISPRS Journal of Photogrammetry and Remote Sensing 116, pp.~1--14.

\bibitem[Hu et al., 2015a]{hu2015transferring}
Hu, F., Xia, G.-S., Hu, J. and Zhang, L., 2015a.
\newblock Transferring deep convolutional neural networks for the scene
  classification of high-resolution remote sensing imagery.
\newblock Remote Sensing 7(11), pp.~14680--14707.

\bibitem[Hu et al., 2015b]{hu2015deep}
Hu, W., Huang, Y., Wei, L., Zhang, F. and Li, H., 2015b.
\newblock Deep convolutional neural networks for hyperspectral image
  classification.
\newblock Journal of Sensors.

\bibitem[Jiang et al., 2018]{jiang2018pointsift}
Jiang, M., Wu, Y. and Lu, C., 2018.
\newblock Pointsift: A sift-like network module for 3d point cloud semantic
  segmentation.
\newblock arXiv preprint arXiv:1807.00652.

\bibitem[Kada and McKinley, 2009]{kada20093d}
Kada, M. and McKinley, L., 2009.
\newblock 3d building reconstruction from lidar based on a cell decomposition
  approach.
\newblock International Archives of Photogrammetry, Remote Sensing and Spatial
  Information Sciences 38(Part 3), pp.~W4.

\bibitem[Kim and Sohn, 2011]{kim2011random}
Kim, H. and Sohn, G., 2011.
\newblock Random forests based multiple classifier system for power-line scene
  classification.
\newblock International Archives of the Photogrammetry, Remote Sensing and
  Spatial Information Sciences 38(5), pp.~W12.

\bibitem[Lalonde et al., 2005]{lalonde2005scale}
Lalonde, J.-F., Unnikrishnan, R., Vandapel, N. and Hebert, M., 2005.
\newblock Scale selection for classification of point-sampled 3d surfaces.
\newblock In: Fifth International Conference on 3-D Digital Imaging and
  Modeling (3DIM'05), IEEE, pp.~285--292.

\bibitem[Lalonde et al., 2006]{lalonde2006natural}
Lalonde, J.-F., Vandapel, N., Huber, D.~F. and Hebert, M., 2006.
\newblock Natural terrain classification using three-dimensional ladar data for
  ground robot mobility.
\newblock Journal of field robotics 23(10), pp.~839--861.

\bibitem[Li et al., 2018a]{li2018so}
Li, J., Chen, B.~M. and Lee, G.~H., 2018a.
\newblock So-net: Self-organizing network for point cloud analysis.
\newblock In: Proceedings of the IEEE Conference on Computer Vision and Pattern
  Recognition, pp.~9397--9406.

\bibitem[Li et al., 2018b]{li2018building}
Li, X., Yao, X. and Fang, Y., 2018b.
\newblock Building-a-nets: Robust building extraction from high-resolution
  remote sensing images with adversarial networks.
\newblock IEEE Journal of Selected Topics in Applied Earth Observations and
  Remote Sensing (99), pp.~1--8.

\bibitem[Li et al., 2018c]{li2018pointcnn}
Li, Y., Bu, R., Sun, M., Wu, W., Di, X. and Chen, B., 2018c.
\newblock Pointcnn: Convolution on x-transformed points.
\newblock In: Advances in Neural Information Processing Systems, pp.~820--830.

\bibitem[Lodha et al., 2007]{lodha2007aerial}
Lodha, S.~K., Fitzpatrick, D.~M. and Helmbold, D.~P., 2007.
\newblock Aerial lidar data classification using adaboost.
\newblock In: Sixth International Conference on 3-D Digital Imaging and
  Modeling (3DIM 2007), IEEE, pp.~435--442.

\bibitem[Maggiori et al., 2016]{maggiori2016convolutional}
Maggiori, E., Tarabalka, Y., Charpiat, G. and Alliez, P., 2016.
\newblock Convolutional neural networks for large-scale remote-sensing image
  classification.
\newblock IEEE Transactions on Geoscience and Remote Sensing 55(2),
  pp.~645--657.

\bibitem[Mallet et al., 2011]{mallet2011relevance}
Mallet, C., Bretar, F., Roux, M., Soergel, U. and Heipke, C., 2011.
\newblock Relevance assessment of full-waveform lidar data for urban area
  classification.
\newblock ISPRS journal of photogrammetry and remote sensing 66(6),
  pp.~S71--S84.

\bibitem[Mongus and {\v{Z}}alik, 2013]{mongus2013computationally}
Mongus, D. and {\v{Z}}alik, B., 2013.
\newblock Computationally efficient method for the generation of a digital
  terrain model from airborne lidar data using connected operators.
\newblock IEEE journal of selected topics in applied earth observations and
  remote sensing 7(1), pp.~340--351.

\bibitem[Munoz et al., 2009]{munoz2009contextual}
Munoz, D., Bagnell, J.~A., Vandapel, N. and Hebert, M., 2009.
\newblock Contextual classification with functional max-margin markov networks.
\newblock In: 2009 IEEE Conference on Computer Vision and Pattern Recognition,
  IEEE, pp.~975--982.

\bibitem[Niemeyer et al., 2012]{niemeyer2012conditional}
Niemeyer, J., Rottensteiner, F. and Soergel, U., 2012.
\newblock Conditional random fields for lidar point cloud classification in
  complex urban areas.
\newblock ISPRS annals of the photogrammetry, remote sensing and spatial
  information sciences 3, pp.~263--268.

\bibitem[Niemeyer et al., 2014]{niemeyer2014contextual}
Niemeyer, J., Rottensteiner, F. and Soergel, U., 2014.
\newblock Contextual classification of lidar data and building object detection
  in urban areas.
\newblock ISPRS journal of photogrammetry and remote sensing 87, pp.~152--165.

\bibitem[Niemeyer et al., 2016]{niemeyer2016hierarchical}
Niemeyer, J., Rottensteiner, F., S{\"o}rgel, U. and Heipke, C., 2016.
\newblock Hierarchical higher order crf for the classification of airborne
  lidar point clouds in urban areas.
\newblock International Archives of the Photogrammetry, Remote Sensing and
  Spatial Information Sciences-ISPRS Archives 41 (2016) 41, pp.~655--662.

\bibitem[Parzen, 1962]{parzen1962}
Parzen, E., 1962.
\newblock On estimation of a probability density function and mode.
\newblock The Annals of Mathematical Statistics 33(3), pp.~1065–1076.

\bibitem[Qi et al., 2017a]{qi2017pointnet}
Qi, C.~R., Su, H., Mo, K. and Guibas, L.~J., 2017a.
\newblock Pointnet: Deep learning on point sets for 3d classification and
  segmentation.
\newblock In: Proceedings of the IEEE Conference on Computer Vision and Pattern
  Recognition, pp.~652--660.

\bibitem[Qi et al., 2017b]{qi2017pointnet++}
Qi, C.~R., Yi, L., Su, H. and Guibas, L.~J., 2017b.
\newblock Pointnet++: Deep hierarchical feature learning on point sets in a
  metric space.
\newblock In: Advances in Neural Information Processing Systems,
  pp.~5099--5108.

\bibitem[Ronneberger et al., 2015]{ronneberger2015u}
Ronneberger, O., Fischer, P. and Brox, T., 2015.
\newblock U-net: Convolutional networks for biomedical image segmentation.
\newblock In: International Conference on Medical image computing and
  computer-assisted intervention, Springer, pp.~234--241.

\bibitem[Shapovalov et al., 2010]{shapovalov2010nonassociative}
Shapovalov, R., Velizhev, E. and Barinova, O., 2010.
\newblock Nonassociative markov networks for 3d point cloud classification.
  the.
\newblock In: International Archives of the Photogrammetry, Remote Sensing and
  Spatial Information Sciences XXXVIII, Part 3A, Citeseer.

\bibitem[Shen et al., 2018]{shen2018mining}
Shen, Y., Feng, C., Yang, Y. and Tian, D., 2018.
\newblock Mining point cloud local structures by kernel correlation and graph
  pooling.
\newblock In: Proceedings of the IEEE conference on computer vision and pattern
  recognition, pp.~4548--4557.

\bibitem[Solberg et al., 2009]{solberg2009mapping}
Solberg, S., Brunner, A., Hanssen, K.~H., Lange, H., N{\ae}sset, E.,
  Rautiainen, M. and Stenberg, P., 2009.
\newblock Mapping lai in a norway spruce forest using airborne laser scanning.
\newblock Remote Sensing of Environment 113(11), pp.~2317--2327.

\bibitem[Su et al., 2015]{su2015multi}
Su, H., Maji, S., Kalogerakis, E. and Learned-Miller, E., 2015.
\newblock Multi-view convolutional neural networks for 3d shape recognition.
\newblock In: Proceedings of the IEEE international conference on computer
  vision, pp.~945--953.

\bibitem[Wang et al., 2018]{wang2018deep}
Wang, S., Suo, S., Ma, W.-C., Pokrovsky, A. and Urtasun, R., 2018.
\newblock Deep parametric continuous convolutional neural networks.
\newblock In: Proceedings of the IEEE Conference on Computer Vision and Pattern
  Recognition, pp.~2589--2597.

\bibitem[Weinmann et al., 2015a]{weinmann2015semantic}
Weinmann, M., Jutzi, B., Hinz, S. and Mallet, C., 2015a.
\newblock Semantic point cloud interpretation based on optimal neighborhoods,
  relevant features and efficient classifiers.
\newblock ISPRS Journal of Photogrammetry and Remote Sensing 105, pp.~286--304.

\bibitem[Weinmann et al., 2015b]{weinmann2015contextual}
Weinmann, M., Schmidt, A., Mallet, C., Hinz, S., Rottensteiner, F. and Jutzi,
  B., 2015b.
\newblock Contextual classification of point cloud data by exploiting
  individual 3d neigbourhoods.
\newblock ISPRS Annals of the Photogrammetry, Remote Sensing and Spatial
  Information Sciences II-3 (2015), Nr. W4 2(W4), pp.~271--278.

\bibitem[Wen et al., 2019]{wen2019directionally}
Wen, C., Yang, L., Peng, L., Li, X. and Chi, T., 2019.
\newblock Directionally constrained fully convolutional neural network for
  airborne lidar point cloud classification.
\newblock arXiv preprint arXiv:1908.06673.

\bibitem[Yang et al., 2017a]{yang2017automated}
Yang, B., Huang, R., Li, J., Tian, M., Dai, W. and Zhong, R., 2017a.
\newblock Automated reconstruction of building lods from airborne lidar point
  clouds using an improved morphological scale space.
\newblock Remote Sensing 9(1), pp.~14.

\bibitem[Yang et al., 2017b]{yang2017convolutional}
Yang, Z., Jiang, W., Xu, B., Zhu, Q., Jiang, S. and Huang, W., 2017b.
\newblock A convolutional neural network-based 3d semantic labeling method for
  als point clouds.
\newblock Remote Sensing 9(9), pp.~936.

\bibitem[Yang et al., 2018]{yang2018segmentation}
Yang, Z., Tan, B., Pei, H. and Jiang, W., 2018.
\newblock Segmentation and multi-scale convolutional neural network-based
  classification of airborne laser scanner data.
\newblock Sensors 18(10), pp.~3347.

\bibitem[Yousefhussien et al., 2018]{yousefhussien2018multi}
Yousefhussien, M., Kelbe, D.~J., Ientilucci, E.~J. and Salvaggio, C., 2018.
\newblock A multi-scale fully convolutional network for semantic labeling of 3d
  point clouds.
\newblock ISPRS journal of photogrammetry and remote sensing 143, pp.~191--204.

\bibitem[Zhan et al., 2017]{zhan2017change}
Zhan, Y., Fu, K., Yan, M., Sun, X., Wang, H. and Qiu, X., 2017.
\newblock Change detection based on deep siamese convolutional network for
  optical aerial images.
\newblock IEEE Geoscience and Remote Sensing Letters 14(10), pp.~1845--1849.

\bibitem[Zhang et al., 2013]{zhang2013svm}
Zhang, J., Lin, X. and Ning, X., 2013.
\newblock Svm-based classification of segmented airborne lidar point clouds in
  urban areas.
\newblock Remote Sensing 5(8), pp.~3749--3775.

\bibitem[Zhao and Popescu, 2009]{zhao2009lidar}
Zhao, K. and Popescu, S., 2009.
\newblock Lidar-based mapping of leaf area index and its use for validating
  globcarbon satellite lai product in a temperate forest of the southern usa.
\newblock Remote Sensing of Environment 113(8), pp.~1628--1645.

\bibitem[Zhao et al., 2018]{zhao2018classifying}
Zhao, R., Pang, M. and Wang, J., 2018.
\newblock Classifying airborne lidar point clouds via deep features learned by
  a multi-scale convolutional neural network.
\newblock International Journal of Geographical Information Science 32(5),
  pp.~960--979.

\end{thebibliography}
\end{spacing}
}



\end{document}